\documentclass{article}
\pdfoutput=1
\usepackage[final]{corl_2017} 

\usepackage[utf8]{inputenc} 
\usepackage[T1]{fontenc}    
\usepackage{hyperref}       
\usepackage{url}            
\usepackage{booktabs}       
\usepackage{amsfonts}       

\usepackage{wrapfig}
\usepackage{amsmath}

\usepackage[pdftex]{graphicx} 
\usepackage{algorithm}
\usepackage{algorithmic}

\title{Path Integral Networks: \\ End-to-End Differentiable Optimal Control}

\author{
  Masashi Okada, Takenobu Aoshima \\
  AI Solutions Center, Panasonic Corporation.\\
  \texttt{\{okada.masashi001,aoshima.takenobu\}@jp.panasonic.com} \\
  \And
  Luca Rigazio \\
  Panasonic Silicon Valley Laboratory, \\
  Panasonic R\&D Company of America \\
  \texttt{luca.rigazio@us.panasonic.com} \\
}

\newcommand{\mcsymbol}[2]{\mathcal{#1}_{#2}}

\begin{document}
\maketitle


\begin{abstract}
In this paper, we introduce Path Integral Networks (PI-Net), a recurrent network representation of the Path Integral optimal control algorithm.
The network includes both system dynamics and cost models, used for optimal control based planning.
PI-Net is fully differentiable, learning both dynamics and cost models end-to-end by back-propagation and stochastic gradient descent. Because of this, PI-Net can learn to plan.
PI-Net has several advantages: it can generalize to unseen states thanks to planning, it can be applied to continuous control tasks, and it allows for a wide variety learning schemes, including imitation and reinforcement learning.
Preliminary experiment results show that PI-Net, trained by imitation learning, can mimic control demonstrations for two simulated problems; a linear system and a pendulum swing-up problem. We also show that PI-Net is able to learn dynamics and cost models latent in the demonstrations.
\end{abstract}



\section{Introduction} \label{sec:intro}
Recently, deep architectures such as convolutional neural networks have been successfully applied to difficult control tasks such as autonomous driving \cite{bojarski2016end}, robotic manipulation \cite{levine2016end} and playing games \cite{mnih2015human}.
In these settings, a deep neural network is typically trained with reinforcement learning or imitation learning to represent a control policy which maps input states to control sequences.
However, as already discussed in \cite{chen2015deepdriving,tamar2016value},
the resulting networks and encoded policies are inherently \textit{reactive}, thus unable to execute \textit{planning} to decide following actions, which may explain poor generalization to new or \textit{unseen} environments.
Conversely, optimal control algorithms utilize specified models of system dynamics and a cost function to predict future states and future cost values. This allows to compute control sequences that minimize expected cost. Stated differently, optimal control executes planning for decision making to provide better generalization.

The main practical challenge of optimal control is specifying system dynamics and cost models. Model-based reinforcement learning \cite{schmidhuber1990line,deisenroth2011pilco} can be used to estimate system dynamics by interacting with the environment.
However in many robotic applications, accurate system identification is difficult.
Furthermore, predefined cost models accurately describing controller goals are required.
Inverse optimal control or inverse reinforcement learning estimates cost models from human demonstrations \cite{ng1999policy, abbeel2004apprenticeship, ziebart2008maximum}, but require perfect knowledge of system dynamics.
Other inverse reinforcement learning methods such as \cite{boularias2011relative, kalakrishnan2013learning,finn2016guided} do not require system dynamics perfect knowledge, however, they limit the policy or cost model to the class of time-varying linear functions.

In this paper, we propose a new approach to deal with these limitations.
The key observation is that control sequences resulting from a specific optimal control algorithm, the path integral control algorithm \cite{williams2017model,williams2016aggressive}, are differentiable with respect to all of the controller internal parameters. The controller itself can thus be represented by a special kind recurrent network, which we call \textit{path integral network} (PI-Net).
The entire network, which includes dynamics and cost models, can then be trained end-to-end using standard back-propagation and stochastic gradient descent with fully specified or approximated cost models and system dynamics.
After training, the network will then execute \textit{planning} by effectively running path integral control utilizing the learned system dynamics and cost model.
Furthermore, the effect of modeling errors in learned dynamics can be mitigated by end-to-end training because cost model could be trained to compensate the errors.

We demonstrate the effectiveness of PI-Net by training the network to imitate optimal controllers of two control tasks: linear system control and pendulum swing-up task.
We also demonstrate that dynamics and cost models, latent in demonstrations, can be adequately extracted through imitation learning.


\section{Path Integral Optimal Control} \label{sec:path_integral_control}
\newcommand{\mbx}{\mathbf{x}}
\newcommand{\mbu}{\mathbf{u}}
Path integral control provides a framework for stochastic optimal control based on Monte-Carlo simulation of multiple trajectories \cite{kappen2005linear}.
This framework has generally been applied to policy improvements for parameterized policies such as dynamic movement primitives \cite{theodorou2010generalized,kalakrishnan2013learning}.
Meanwhile in this paper, we focus on a state-of-the-art path-integral optimal control algorithm \cite{williams2017model,williams2016aggressive} developed for model predictive control (MPC; a.k.a.~receding horizon control).
In the rest of this section, we briefly review this path integral optimal algorithm.

Let $\mbx_{t_{i}} \in \mathbb{R}^{n}$ denote the state of a dynamical system at discrete time $t_{i}$,
$\mbu_{t_{i}} \in \mathbb{R}^{m}$ denotes a control input for the system.
This paper supposes that the dynamics take the following form:
\begin{equation}
  \mbx_{t_{i+1}} = f(\mbx_{t_i}, \mbu_{t_i} + \delta\mbu_{t_i}),
  \label{eqn:dynamics}
\end{equation}
where $f: \mathbb{R}^{n}\rightarrow \mathbb{R}^{n}$ is a dynamics function and $\delta\mbu_{t_i} \in \mathbb{R}^{m}$ is a Gaussian noise vector with deviation $\sigma$.
The stochastic optimal control problem here is to find the optimal control sequence $\{\mbu^{*}_{t_i}\}_{i=0}^{N-1}$ which minimizes the expected value of trajectory cost function $S(\tau_{t_0})$:
\begin{equation}
  J = \mathbb{E}\left[S(\tau_{t_0})\right] =
    \mathbb{E}\left[\phi(\mbx_{t_{N}}) +
    \sum^{N-1}_{i=0}\left(q(\mbx_{t_{i}}) +
    \frac{1}{2}\mbu_{t_{i}}^{T}R\mbu_{t_{i}}
    \right)
    \right], \label{eqn:soc_obj}
\end{equation}
where $\mathbb{E}[\cdot]$ denotes an expectation values with respect to trajectories $\tau_{t_0} = \{\mbx_{t_{0}}, \mbu_{t_{0}}, \mbx_{t_{1}}, \cdots, \mbx_{t_{N-1}}\}$ by Eq.~(\ref{eqn:dynamics}).
$\phi: \mathbb{R}^n \rightarrow \mathbb{R}$ and $q: \mathbb{R}^n \rightarrow \mathbb{R}$ are respectively terminal- and running-cost; they are arbitrary state-dependent functions.
$R \in \mathbb{R}^{m \times m}$ is a positive definite weight matrix of the quadratic control cost.
In ~\cite{williams2017model}, a path integral algorithm has been derived to solve this optimization problem, which iteratively improves $\{\mbu_{t_i}\}$%
\footnote{In the rest of this paper, \{$\mbu_{t_i}\}$ represents a sequence \{$\mbu_{t_i}\}^{N-1}_{i=0}$.}
to $\{\mbu_{t_i}^{*}\}$ by the following update law:

\begin{equation}
  \mbu^{*}_{t_i} \leftarrow \mbu_{t_i} +
    \mathbb{E}\left[\exp(-\tilde{S}(\tau_{t_i})/\lambda)\cdot \delta\mbu_{t_i}\right] \bigg/
    \mathbb{E}\left[\exp(-\tilde{S}(\tau_{t_i})/\lambda)\right], \label{eqn:update_law}
\end{equation}
where $\tilde{S}(\tau_{t_i})$ is a modified trajectory cost function:
\begin{equation}
  \tilde{S}(\tau_{t_i}) = \phi(\mbx_{t_N}) + \sum^{N-1}_{j=i} \tilde{q}(\mbx_{t_{j}}, \mbu_{t_{j}}, \delta\mbu_{t_{j}}), \label{eqn:cost-to-go}
\end{equation}
\begin{equation}
  \tilde{q}(\mbx, \mbu, \delta\mbu) = q(\mbx) + \frac{1}{2}\mbu^{T}R\mbu +
    \frac{1-\nu^{-1}}{2}\delta\mbu^{T}R\delta\mbu + \mbu^{T}R\delta\mbu. \label{eqn:modified_cost}
\end{equation}
Eq.~(\ref{eqn:update_law}) can be implemented on digital computers
by approximating the expectation value with the Monte Carlo method as shown in Alg.~\ref{alg:pi}.
\begin{wrapfigure}{R}{0.5\textwidth}
  \begin{minipage}{0.5\textwidth}
  	\vspace*{-5mm}
    \begin{algorithm}[H]
      \caption{Path Integral Optimal Control} \label{alg:pi}
      \begin{algorithmic}[1]
        \INPUT $K$, $N$: Num.~of trajectories \& timesteps \\
        $\mbx_{t_{0}}$ : State \\
        $\{\mbu_{t_{i}}\}$: Initial control sequence \\
        $\{\delta\mbu^{(k)}_{t_{i}}\}$: Gaussian noise \\
        $f, q, \phi, R$: Dynamics and cost models \\
        $\lambda, \nu$: Hyper-parameters
        \OUTPUT $\{\mbu^{*}_{t_{i}}\}$: Improved control sequence \\
		\FOR{$k \leftarrow 0$ \textbf{to} $K-1$}
		\STATE $\mbx^{(k)}_{t_{0}} \leftarrow \mbx_{t_{0}}$
		\FOR{$i \leftarrow 0$ \textbf{to} $N-1$}
		\STATE $q^{(k)}_{t_{i}} \leftarrow \tilde{q}(\mbx_{t_i}^{(k)}, \mbu_{t_i}, \delta\mbu^{(k)}_{t_i})$
		\STATE $\mbx^{(k)}_{t_{i+1}} \leftarrow f\left(\mbx^{(k)}_{t_{i}}, \mbu_{t_{i}} + \delta\mbu^{(k)}_{t_{i}}\right)$
		\ENDFOR
		\STATE $q^{(k)}_{t_{N}} \leftarrow \phi(\mbx^{(k)}_{t_{N}})$
		\ENDFOR
		\FOR{$k \leftarrow 0$ \textbf{to} $K-1$}
		\FOR{$i \leftarrow 0$ \textbf{to} $N$}
		\STATE $\tilde{S}^{(k)}_{\tau_{t_{i}}} \leftarrow \sum^{N}_{j=i}q^{(k)}_{t_{N}}$
		\ENDFOR
		\ENDFOR

		\FOR{$i \leftarrow 0$ \textbf{to} $N-1$}
		\STATE $\mbu^{*}_{t_i} \leftarrow \mbu_{t_i} + \frac{\sum^{K-1}_{k=0}\left[\exp\left(-\tilde{S}^{(k)}_{\tau_{t_{i}}}/\lambda\right)\cdot\delta\mathbf{u}^{(k)}_{t+\tau}\right]}{\sum^{K-1}_{k=0}\left[\exp\left(-\tilde{S}^{(k)}_{\tau_{t_{i}}}/\lambda\right)\right]}$
		\ENDFOR
      \end{algorithmic}
    \end{algorithm}
    \vspace*{-10mm}
  \end{minipage}
\end{wrapfigure}

Different from other general optimal control algorithms, such as iterative linear quadratic regulator (iLQR) \cite{todorov2005generalized},  path integral optimal control does not require first or second-order approximation of the dynamics and a quadratic approximation of the cost model, naturally allowing for non-linear system dynamics and cost models. This flexibility allows us to use general function approximators, such as neural networks, to represent dynamics and cost models in the most general possible form.


\section{Path Integral Networks} \label{sec:path_integral_networks}
\begin{figure*}[t]
	\centering
	\includegraphics[width=\textwidth]{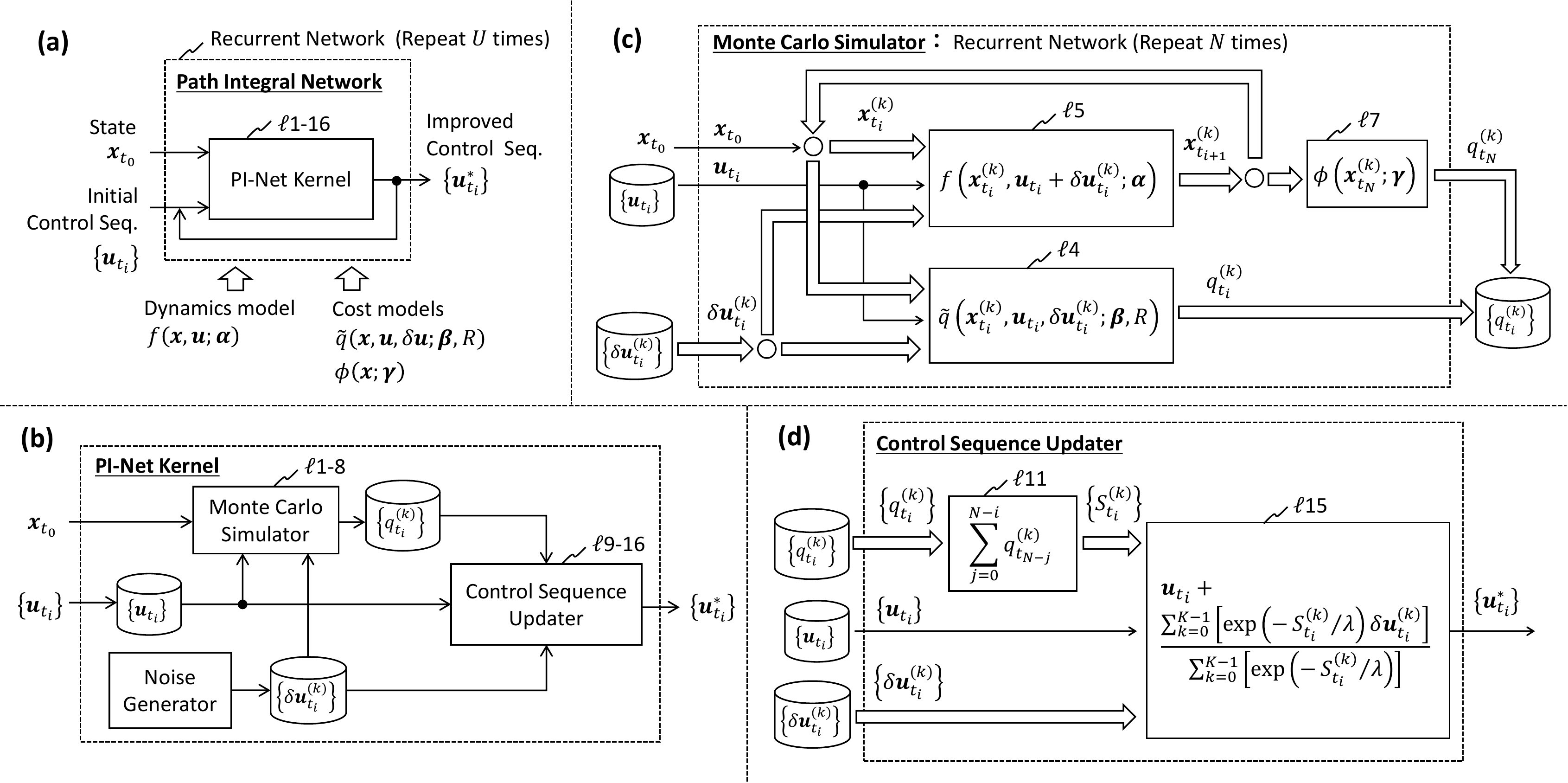}
    \vspace*{-5mm}
	\caption{
  Architecture of PI-Net.
  Labels with `$\ell$' indicate corresponding line numbers in Alg.~\ref{alg:pi}.
  Block arrows in (c,d) indicate multiple signal flow with respect to $K$ trajectories.
  } \label{fig:pi_arch}
\end{figure*}
\subsection{Architecture}
We illustrate the architecture of PI-Net in Figs.~\ref{fig:pi_arch} (a)-(d).
The architecture encodes Alg.~\ref{alg:pi} as a fully differentiable recurrent network representation.
Namely, the forward pass of this network completely imitates the iterative execution of Alg.~\ref{alg:pi}.

\newcommand{\zw}{\hspace{0.25cm}}
\textbf{Top-level architecture} of PI-Net is illustrated in Fig.~\ref{fig:pi_arch}~(a).
This network processes input current state $\mbx_{t_0}$ and initial control sequence $\{\mbu_{t_i}\}$ to output a control sequence $\{\mbu^{*}_{t_i}\}$ improved by the path integral algorithm.
In order to execute the algorithm, dynamics model $f$ and cost models $\tilde{q}, \phi$ are also given and embedded into the network.
We suppose $f$, $\tilde{q}$, $\phi$ are respectively parameterized by ${\boldsymbol \alpha}$, $({\boldsymbol \beta}, R)$\footnote{${\boldsymbol \beta}$ is a parameter for a state-dependent cost model $q$; see Eqs.~(\ref{eqn:soc_obj}, \ref{eqn:modified_cost})} and ${\boldsymbol \gamma}$, which are what we intend to train.
We remark that PI-Net architecture allows to use both approximated parameterized models (e.g., neural networks) or explicit models for both system dynamics and cost models.
In the network, \textit{PI-Net Kernel} module is recurrently connected, representing iterative execution Alg.~\ref{alg:pi}. The number of recurrence $U$ is set to be sufficiently large for convergence.

\textbf{PI-Net Kernel} in Fig.~\ref{fig:pi_arch}~(b) contains three modules: \textit{Noise Generator}, \textit{Monte-Carlo Simulator} and \textit{Control Sequence Updater}.
First, the \textit{Noise Generator} procures $K\times N$ Gaussian noise vectors $\delta \mbu_{t_i}^{(k)}$ sampled from $\mathcal{N}(0, \sigma)$.
Then the noise vectors are input to \textit{Monte-Carlo Simulator} along with $\mbx_{t_0}$ and $\{\mbu_{t_i}\}$, which estimates running- and terminal-cost values (denoted as $q_{t_i}^{(k)}$) of different $K$ trajectories.
Finally, the estimated cost values are fed into the \textit{Control Sequence Updater} to improve the initial control sequence.

\textbf{Monte Carlo Simulator} in Fig.~\ref{fig:pi_arch}~(c) contains the system dynamics $f$ and cost models $\tilde{q}, \phi$, which are responsible to predict future states and costs over $K$ trajectories. The simulations of $K$ trajectories are conducted in parallel.
The prediction sequence over time horizon is realized by network recurrence. In the $i$-th iteration ($i \in \{0, 1, \cdots, N-1\}$), current $K$ states $\mbx^{(k)}_{t_i}$ and perturbated controls $\mbu_{t_i} + \delta\mbu^{(k)}_{t_i}$ are input to the dynamics model $f$ to predicted next $K$ states $\mbx^{(k)}_{t_{i+1}}$.
The predicted states $\mbx^{(k)}_{t_{i+1}}$ are feedbacked for next iteration.
While in the $i$-th iteration, the above inputs are also fed into the cost model $\tilde{q}$ to compute running-cost values.
Only in the last iteration ($i=N-1$), predicted terminal-states $\mbx^{(k)}_{t_{N}}$ are input to the cost model $\phi$ to compute terminal-cost values.
This module is a recurrent network in a recurrent network, making entire PI-Net a \textit{nested} or \textit{double-looped} recurrent network.

\textbf{Control Sequence Updater} in Fig.~\ref{fig:pi_arch}~(d) update input control sequence based on the equations appeared in $\ell$9--$\ell$16 in Alg.~\ref{alg:pi}.
Since all equations in the loops can be computed in parallel, no recurrence is needed for this module.

\subsection{Learning schemes}
We remark that all the nodes in the computational graph of PI-Net are differentiable. We can therefore employ the chain rule to differentiate the network end-to-end, concluding that PI-Net is \textit{fully differentiable}.
If an objective function with respect to the network control output, denoted as $\mcsymbol{L}{ctrl}$, is defined, then we can differentiate the function with the internal parameters (${\boldsymbol \alpha}, {\boldsymbol \beta}, R, {\boldsymbol \gamma}$). Therefore, we can tune the parameters by optimizing the objective function with gradient descent methods.
In other words, we can train internal dynamics $f$ and/or cost models $q, \phi, R$ end-to-end through the optimization.
For the optimization, we can re-use all the standard Deep Learning machinery, including back-propagation and stochastic gradient descent, and a variety of Deep Learning frameworks.
We implemented PI-Net with TensorFlow \cite{dean2015tensorflow}.
Interestingly, all elemental operations of PI-Net can be described as TensorFlow nodes, allowing to utilize automatic differentiation.

A general use case of PI-Net is imitation learning to learn dynamics and cost models latent in experts' demonstrations.
Let us consider an open loop control setting and suppose that a dataset $\mathcal{D} \ni (\mbx^{\star}_{t_0}, \{\mbu^{\star}_{t_i}\})$ is available; $\mbx^{\star}_{t_0}$ is a state observation and $\{\mbu^{\star}_{t_i}\}$ is a corresponding control sequence generated by an expert.
In this case, we can supervisedly train the network by optimizing $\mcsymbol{L}{ctrl}$, i.e., the errors between the expert demonstration $\{\mbu^{\star}_{t_i}\}$ and the network output $\{\mbu^{*}_{t_i}\}$.
For closed loop  infinite time horizon control setting, the network can be trained as an MPC controller. If we have a trajectory by an expert $\{\mbx^{\star}_{t_{0}}, \mbu^{\star}_{t_{0}}, \mbx^{\star}_{t_{1}}, \cdots, \}$, we can construct a dataset $\mathcal{D} \ni (\mbx^{\star}_{t_i}, \mbu^{\star}_{t_i})$ and then optimize the estimation errors between the expert control $\mbu^{\star}_{t_i}$ and the first value of output control sequence output.
If \textit{sparse reward function} is available, reinforcement learning could be introduced to train PI-Net.
The objective function here is \textit{expected return} which can be optimized by policy gradient methods such as REINFORCE \cite{williams1992simple}.

\textbf{Loss functions} \zw
In addition to $\mcsymbol{L}{ctrl}$, we can append other loss functions to make training faster and more stable.
In an MPC scenario, we can construct a dataset in another form $\mathcal{D} \ni (\mbx^{\star}_{t_i}, \mbu^{\star}_{t_i}, \mbx^{\star}_{t_{i+1}})$. In this case, a loss function $\mcsymbol{L}{dyn}$ with respect to internal dynamics output can be introduced; i.e.,~state prediction errors between $f(\mbx^{\star}_{t_i}, \mbu^{\star}_{t_i})$ and $\mbx^{\star}_{t_{i+1}}$.
Furthermore, we can employ loss functions $\mcsymbol{L}{cost}$ regarding cost models.
In many cases on control domains, we know goal states $\mbx_{g}$ in prior and we can assume cost models $q, \phi$ have optimum points at $\mbx_g$.
Therefore, loss functions, which penalize conditions of $q(\mbx_g) > q(\mbx)$, can be employed to help the cost models have such property. This is a useful approach when we utilize highly expressive approximators (e.g., neural networks) to cost models.
In the later experiments, mean squared error (MSE) was used for $\mcsymbol{L}{ctrl,dyn}$.
$\mcsymbol{L}{cost}$ was defined as $\varphi(q(\mbx_g) - q(\mbx))$, where $\varphi$ is the ramp function.
The sum of these losses can be jointly optimized in a single optimization loop. Of course, dynamics model can be pre-trained independently by optimizing $\mcsymbol{L}{dyn}$.

\subsection{Discussion of computational complexity}
In order to conduct back-propagation, we must store all values fed into computational graph nodes during preceding forward pass.
Let $B$ be a mini-batch size, then internal dynamics $f$ and running-cost model $\tilde{q}$ in PI-Net are evaluated $U \times N \times K \times B$ times during the forward pass;
this value can grow very fast, making optimization memory hungry.
For instance, the experiments of Sect.~\ref{sec:exp} used over 100GB of RAM, forcing us to train on CPU instead of GPU. 

The complexity can be alleviated by data parallel approach, in which a mini-batch is divided and processed in parallel with distributed computers. Therefore, we can reduce the batch size $B$ processed on a single computer.
Another possible approach is to reduce $U$; the recurrence number of the \textit{PI-Net Kernel} module.
In the experiment, initial control sequence is filled with a constant value (i.e., zero) and $U$ is set to be large enough (e.g., $U=200$).
In our preliminary experiment, we found that inputting desired output (i.e.,~demonstrations) as initial sequences and training PI-Net with small $U$ did not work; the trained PI-Net just passed through the initial sequence, resulting in poor generalization performance.
In the future, a scheme to determine good initial sequences, which reduces $U$ while achieving good generalization, must be established.

Note that the memory problem is valid only during training phase because the network does not need to store input values during control phase.
In addition, the mini-batch size is obviously $B=1$ in that phase.
Further in MPC scenarios, we can employ \textit{warm start settings} to reduce $U$, under which output control sequences are re-used as initial sequences at next timestep.
For instance in \cite{williams2016aggressive, williams2017model}, real-time path integral control has been realized by utilizing GPU parallelization.


\section{Related Work} \label{sec:related_works}
Conceptually PI-Net is inspired by the \textit{value iteration network} (VIN) \cite{tamar2016value}, a differentiable network representation of the value iteration algorithm designed to train internal state-transition and reward models end-to-end.
The main difference between VIN and PI-Net lies in the underlying algorithms: the value iteration, generally used for discrete Markov Decision Process (MDP), or path integral optimal control, which allows for continuous control.
In \cite{tamar2016value}, VIN was applied to 2D navigation task, where 2D space was discretized to grid map and a reward function was defined on the discretized map. In addition, action space was defined as eight directions to forward. The experiment showed that adequately estimated reward map can be utilized to navigate an agent to goal states by not only discrete control but also continuous control.
Let us consider a more complex 2D navigation task on continuous control, in which velocity must be taken into reward function%
\footnote{Such as a task to control a mass point to trace a fixed track while forwarding it as fast as possible.}.
In order to design such the reward function with VIN, 4D state space (position and velocity) and 2D control space (vertical and horizontal accelerations) must be discretized.
This kind of discretization could cause combinatorial explosion especially for higher dimensional tasks.

Generally used optimal controller, linear quadratic regulator (LQR), is also differentiable and Ref.~\cite{tamar2016learning} employs this insight to \textit{re-shape} original cost models to improve short-term MPC performance. 
The main advantage of the path integral control over (iterative-)LQR is that we do not require a linear and quadratic approximation of non-linear dynamics and cost model. In order to differentiate iLQR with non-linear models by back-propagation, we must iteratively differentiate the functions during preceding forward pass, making the backward pass very complicated.

Policy Improvement with Path Integrals \cite{theodorou2010generalized} and Inverse Path Integral Inverse Reinforcement Learning  \cite{kalakrishnan2013learning} are policy search approaches based on the path integral control framework, which train a parameterized control policy via reinforcement learning and imitation learning, respectively.
These methods have been succeeded to train policies for complex robotics tasks, however, they assume trajectory-centric policy representation such as dynamic movement primitives \cite{ijspeert2002movement};
such the policy is less generalizable for \textit{unseen} settings (e.g., different initial states).

Since PI-Net is a policy representation of optimal control, trainable end-to-end by standard back-propagation, a wide variety of learning to control approaches may be applied, including:

\textbf{Imitation learning} \zw \textsc{DAgger} \cite{ross2011reduction} and PLATO \cite{kahn2016plato},

\textbf{Reinforcement learning} \zw Deep Deterministic Policy Gradient \cite{lillicrap2015continuous}, A3C \cite{mnih2016asynchronous},
Trust Region Optimization \cite{schulman2015trust},  Guided Policy Search \cite{levine2016end} and Path Integral Guided Policy Search \cite{chebotar2016path}.


\section{Experiments} \label{sec:exp}
We conducted experiments to validate the viability of PI-Net.
These experiments are meant to test if the network can effectively learn policies in an imitation learning setting.
We did this by supplying demonstrations generated by general optimal control algorithms, with known dynamics and cost models (termed \textit{teacher models} in the rest of this paper).
Ultimately, in real application scenarios, demonstrations may be provided by human experts.

\subsection{Linear system}
The objective of this experiment is to validate that PI-Net is trainable and it can jointly learn dynamics and cost models latent in demonstrations.

\textbf{Demonstrations} \zw
Teacher models in this experiment were linear dynamics, $f^{\star}(\mbx, \mbu) = F^{\star}\mbx + G^{\star}\mbu$, and quadratic cost model, $q^{\star}(\mbx) = \phi^{\star}(\mbx) = \mathbf{x}^{T}Q^{\star}\mathbf{x} / 2$, where $\mathbf{x} \in \mathbb{R}^{4}$, $\mathbf{u} \in \mathbb{R}^{2}$,
$F^{\star} \in \mathbb{R}^{4 \times 4}$, $G^{\star} \in \mathbb{R}^{4 \times 2}$,
$Q^{\star} \in \mathbb{R}^{4 \times 4}$, $R^{\star} \in \mathbb{R}^{2 \times 2}$.
Starting from the fixed set of parameters, we produced training and test dataset $\mcsymbol{D}{train}$, $\mcsymbol{D}{test}$, which take the from:
$\mathcal{D} \ni (\mbx^{\star}_{t_0}, \{\mbu^{\star}_{t_i}\})$.
LQR was used to obtain references $\{\mathbf{u}^{\star}_{t_i}\}$ from randomly generated state vectors $\mbx^{\star}_{t_0}$. The size of each dataset was $|\mcsymbol{D}{train}|=950$ and $|\mcsymbol{D}{test}|=50$.

\textbf{PI-Net settings} \zw
Internal dynamics and cost models were also linear and quadratic form whose initial parameters were different from the teacher models'.
PI-Net was supervisedly trained by optimizing $\mcsymbol{L}{ctrl}$.
We did not use $\mcsymbol{L}{dyn}$ and $\mcsymbol{L}{cost}$ in this experiment.

\begin{figure}[t]
    \centering
    \includegraphics[width=0.95\textwidth]{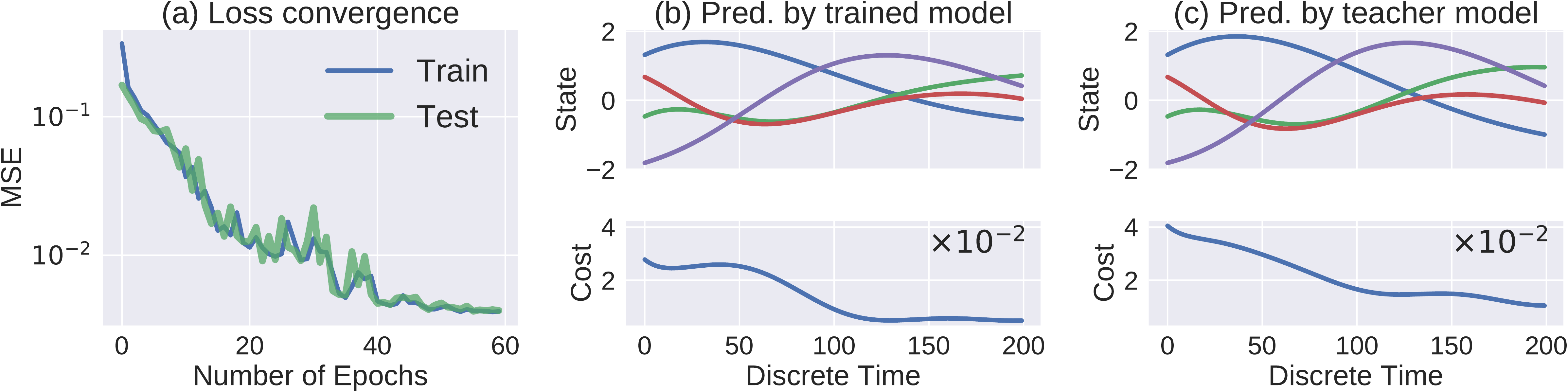}
    \caption{Results in linear system experiment.} \label{fig:exp_linear_model_analysis}
\end{figure}

\textbf{Results} \zw
Fig.~\ref{fig:exp_linear_model_analysis} shows the results of this experiments.
Fig.~\ref{fig:exp_linear_model_analysis}~(a) illustrates loss during training epochs, showing good convergence to a lower fixed point. This validates that PI-Net was indeed training well and the trained policy generalized well to test samples.
Fig.~\ref{fig:exp_linear_model_analysis}~(b, c) exemplifies state and cost trajectories predicted by trained dynamics and cost models, which were generated by feeding certain initial state and corresponding optimal control sequence into the models. Fig.~\ref{fig:exp_linear_model_analysis}~(c) are trajectories by the teacher models.
State trajectories in Fig.~\ref{fig:exp_linear_model_analysis}(b, c) approximate each other, indicating the internal dynamics model learned to approximate the teacher dynamics.
It is well-known that different cost functions could result in same controls \cite{ziebart2008maximum}, and indeed cost trajectories in Fig.~\ref{fig:exp_linear_model_analysis}(b, c) seem not similar.
However, this would not be a problem as long as a learned controller is well generalized to unseen state inputs.

\subsection{Pendulum swing up}
Next, we tried to imitate demonstrations generated from non-linear dynamics and non-quadratic cost models while validating the PI-Net applicability to MPC tasks.
We also compared PI-Net with VIN.

\textbf{Demonstrations} \zw
The experiment focuses on the classical inverted pendulum swing up task \cite{wang1996approach}.
Teacher cost models were $q^{\star}(\theta, \dot{\theta}) = \phi^{\star}(\theta, \dot{\theta}) = (1 + \cos\theta)^{2} + \dot{\theta}^{2},\ R^{\star} = 5$,
where $\theta$ is pendulum angle and $\dot{\theta}$ is angular velocity.
Under this model assumptions, we firstly conducted 40s MPC simulations with iLQR to generate trajectories.
We totally generated $50$ trajectories and then produced $\mcsymbol{D}{train} \ni (\mbx^{\star}_{t_i}, u^{\star}_{t_i}, \mbx^{\star}_{t_{i+1}})$, where control $u$ is torque to actuate a pendulum.
We also generated $10$ trajectories for $\mcsymbol{D}{test}$.

\textbf{PI-Net settings} \zw
We used a more general modeling scheme where internal dynamics and cost models were represented by neural networks, both of which had one hidden layer. The number of hidden nodes was 12 for the dynamics model, and 24 for the cost model.
First, we pre-trained the internal dynamics independently by optimizing $\mcsymbol{L}{dyn}$ and then the entire network was trained by optimizing $\mcsymbol{L}{ctrl} + 10^{-3}\cdot\mcsymbol{L}{cost}$. In this final optimization, the dynamics model was freezed to focus on cost learning.
Goal states used to define $\mcsymbol{L}{cost}$ were $\mbx_{g} = (\theta, \dot{\theta}) = (\pm \pi, 0)$.
We prepared a model variant, termed freezed PI-Net, whose internal dynamics was the above-mentioned pre-trained one and cost model was teacher model as is.
The freezed PI-Net was not trained end-to-end.

\textbf{VIN settings} \zw
VIN was designed to have 2D inputs for continuous states and 1D output for continuous control.
In order to define a reward map embedded in VIN, we discretized 2D continuous state to $31 \times 31$ grid map. This map is cylindrical because angle-axis is cyclic.
We also discretized 1D control space to 7 actions, each of which corresponds to different torque.
We denote the reward map as $\mathcal{R}(s, a) \in \mathbb{R}^{31 \times 31 \times 7}$, where $s$ and $a$ respectively denote discrete state and action.
The reward map can be decomposed as the sum of $\mathcal{R}_{1}(s) \in \mathbb{R}^{31 \times 31 \times 1}$ and $\mathcal{R}_{2}(a) \in \mathbb{R}^{1 \times 1 \times 7}$.
Original VIN employed convolutional neural networks (CNNs) to represent state transition kernels.
However, this representation implies that state transition probability $\mathcal{P}(s'|s,a)$ can be simplified to $\mathcal{P}(s'-s|a)$%
\footnote{
Under this supposition, probability of relative state transition, with respect to a certain action, is invariable.
}.
Since this supposition is invalid for the pendulum system%
\footnote{
See the time evolution equation of this system, $\ddot{\theta} = -\sin\theta + k\cdot u$ ($k$: model paramter); relative time-variation of angular velocity $\dot{\theta}$, with respect to certain torque $u$, varies with pendulum angle $\theta$.
}
, we alternatively employed locally connected neural networks (LCNs; i.e.,~CNNs without weight sharing) \cite{yentis1996vlsi}.
We also prepared a CNN-based VIN for comparison.
The embedded reward maps and transition kernels were trained end-to-end by optimizing $\mcsymbol{L}{ctrl}$.

\begin{figure}[t]
	\centering
  \includegraphics[height=3.25cm]{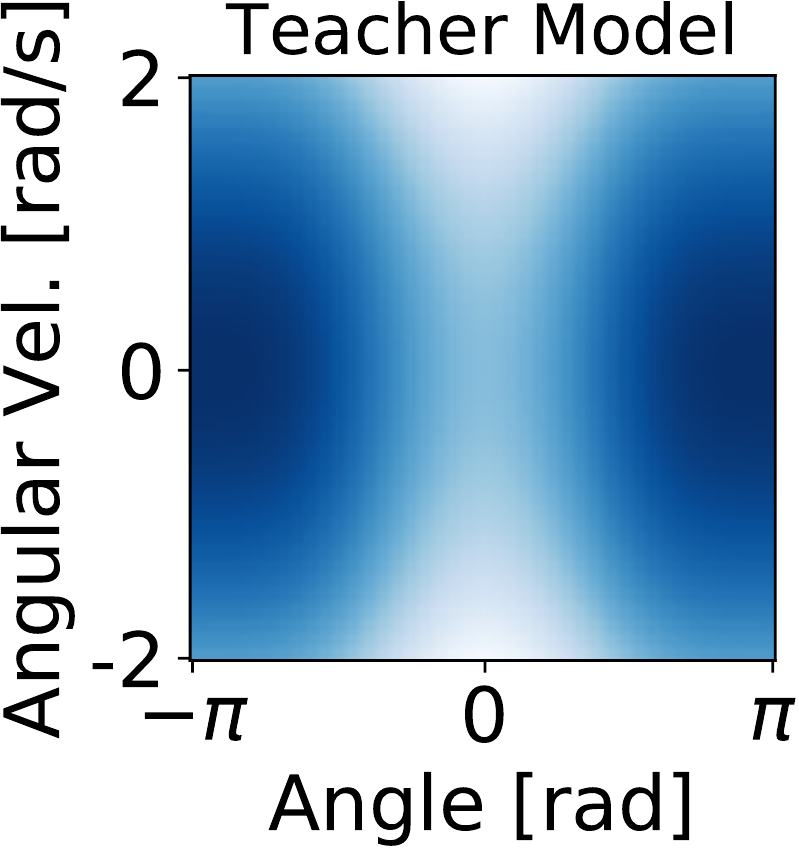}
  \includegraphics[height=3.25cm]{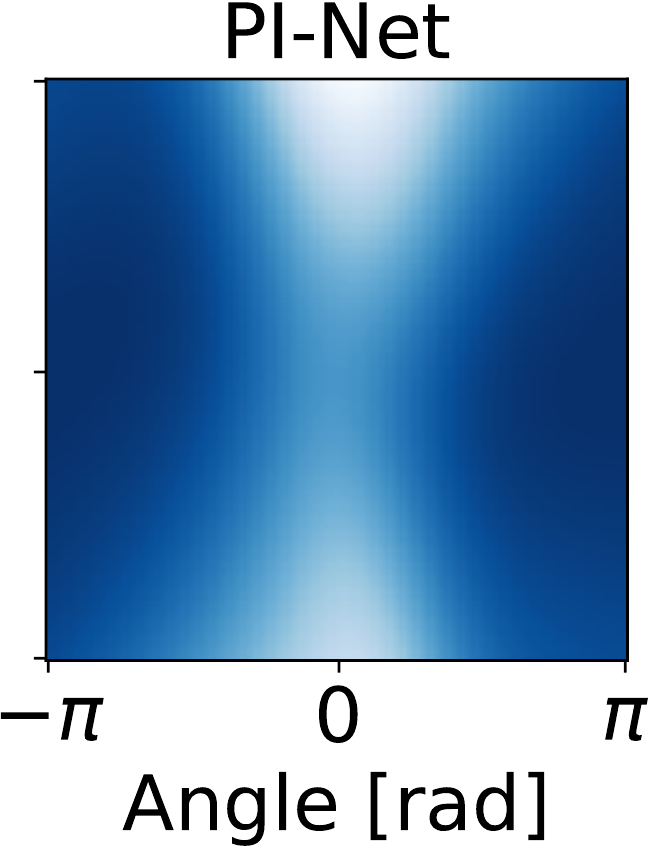}
  \includegraphics[height=3.25cm]{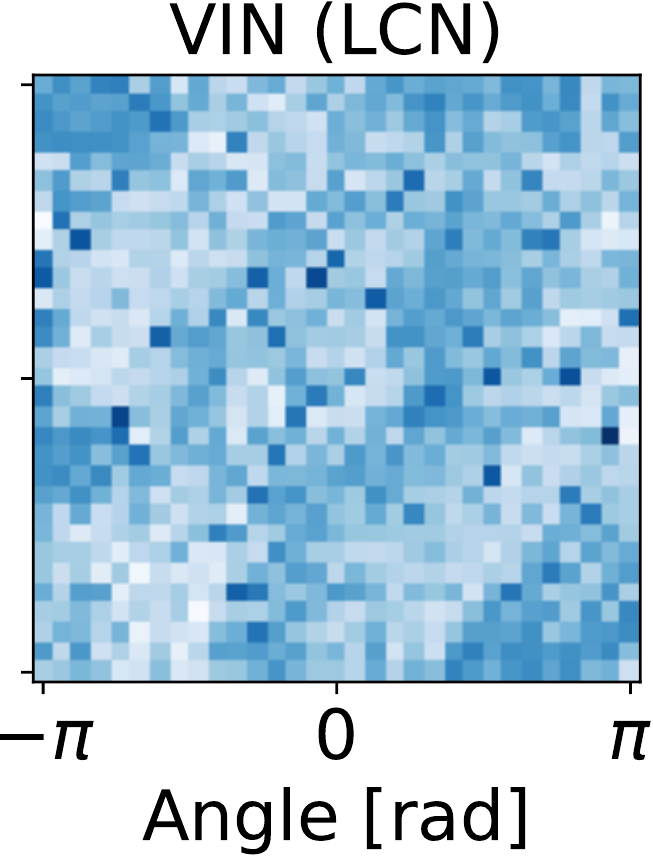}
  \includegraphics[height=3.25cm]{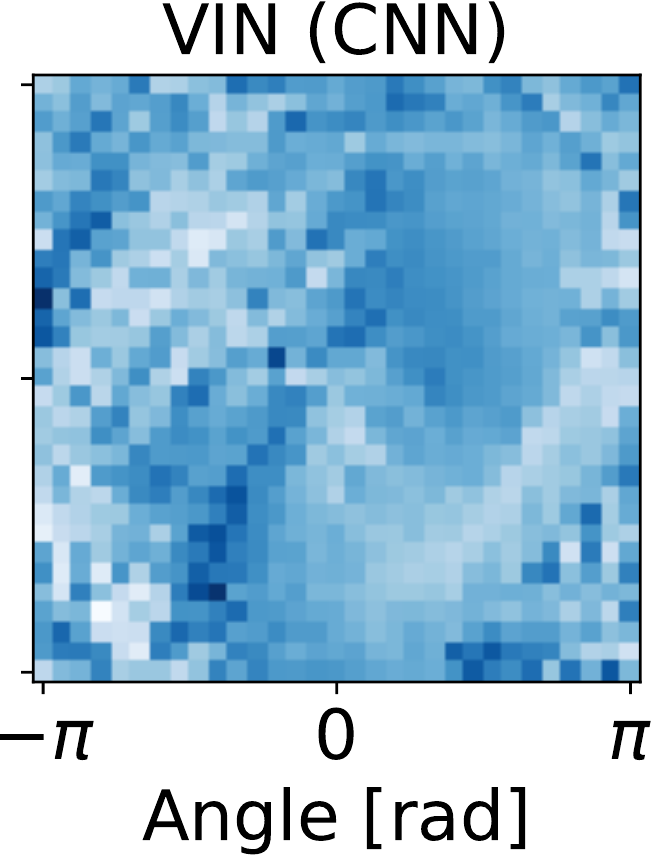}
  \caption{Cost/Reward maps. From left to right, the teacher cost $q^{\star}$, neural cost $q$ trained by PI-Net, and reward maps $\mathcal{R}_{1}$ trained by the LCN- and CNN-based VINs. Dense color indicates low cost (or high reward).}
  \label{fig:pendulum_cost}
\end{figure}

\textbf{Results} \zw
The results of training and MPC simulations with trained controllers are summarized in Table~\ref{tab:pendulum}.
In the simulations, we observed success rates of task completion (defined as keeping the pendulum standing up more than 5s) and trajectory cost $S(\tau)$ calculated by the teacher cost.
For each network, ten 60-second simulations were conducted starting from different initial states.
In the table, freezed PI-Net showed less generalization performance although it was equipped with the teacher cost.
This degradation might result from the modeling errors of the learned dynamics.
On the other hand, trained PI-Net achieved the best performance both on generalization and control, suggesting that adequate cost model was trained to imitate demonstrations while compensating the dynamics errors.
Fig.~\ref{fig:pendulum_cost} illustrates visualized cost models where the cost map of the trained PI-Net resembles the teacher model well.
The reason of VIN failures must result from modeling difficulty on continuous control tasks.
Fine discretization of state and action space would be necessary for good approximation; however, this results in the explosion of parameters to be trained, making optimizations difficult.
CNN-representation of transition kernels would not work because this is very rough approximation for the most of control systems.
Therefore, one can conclude that the use of PI-Net would be more reasonable on continuous control because of the VIN modeling difficulty.
\begin{table}[t]
  \centering
  \caption{Training and simulation results on pendulum swing-up task. Trajectory cost shown here is the average of 10 trajectories.} \label{tab:pendulum}
  \begin{tabular}{cccccc} \hline
    & \textbf{MSE for} & \textbf{MSE for} & \textbf{Success} & \textbf{Traj.~Cost} & \textbf{\# trainable} \\
    \textbf{Controllers} & {\boldmath $\mcsymbol{D}{train}$} & {\boldmath $\mcsymbol{D}{test}$} & \textbf{Rate} & {\boldmath ${S(\tau)}$} & \textbf{params} \\\hline
    Expert & N/A & N/A & 100\% & ${404.63}$ & N/A \\\hline 
    Trained PI-Net & ${2.22 \times 10^{-3}}$ & $\mathbf{1.65 \times 10^{-3}}$ & $\mathbf{100\%}$ & $\mathbf{429.69}$ & 242 \\ 
	  Freezed PI-Net & $\mathbf{1.91 \times 10^{-3}}$ & ${5.73 \times 10^{-3}}$ & ${100\%}$ & 982.22 & 49 \\   
    VIN (LCN) & ${6.44\times 10^{-3}}$ & ${6.89\times 10^{-3}}$ & 0\% & $2409.29$ & 330,768 \\
    VIN (CNN) & ${4.45\times 10^{-3}}$ & ${4.72\times 10^{-3}}$ & 0\% & $1280.62$ & 1,488 \\\hline
  \end{tabular}
\end{table}


\section{Conclusion}
In this paper, we introduced path integral networks, a fully differentiable end-to-end trainable  representation of the path integral optimal control algorithm, which allows for optimal continuous control.
Because PI-Net is fully differentiable, it can rely on powerful Deep Learning machinery to efficiently estimate in high dimensional parameters spaces from large data-sets. To the best of our knowledge, PI-Net is the first end-to-end trainable differentiable optimal controller directly applicable to continuous control domains.

PI-Net architecture is highly flexible, allowing to specify system dynamics and cost models in an explicit way, by analytic models, or in an approximate way by deep neural networks. Parameters are then jointly estimated end-to-end, in a variety of settings, including imitation learning and reinforcement learning. This may be very useful for non-linear continuous control scenarios, such as the ``pixel to torques'' scenario, and in situations where it's difficult to fully specify system dynamics or cost models. We postulate this architecture may allow to train approximate system dynamics and cost models in such complex scenarios while still carrying over the advantages of optimal control from the underlying path integral optimal control. We show promising initial results in an imitation learning setting, comparing against optimal control algorithms with linear and non-linear system dynamics.
Future works include a detailed comparison to other baselines, including other learn-to-control methods as well as experiments in high-dimensional settings.
To tackle high-dimensional problems and to accelerate convergence we plan to combine PI-Net with other powerful methods, such as guided cost learning \cite{finn2016guided} and trust region policy optimization \cite{schulman2015trust}.

\newpage
\bibliography{corl}

\newpage
\appendix
\section{Supplements of Experiments}
\subsection{Common settings}
We used RMSProp for optimization.
Initial learning rate was set to be $10^{-3}$ and the rate was decayed by a factor of 2 when no improvement of loss function was observed for five epochs.
We set the hyper-parameters appeared in the path integral algorithm \cite{williams2017model} as $\lambda = 0.01$, $\nu=1500$.

\subsection{Linear System}
\paragraph{Demonstrations}
Dynamics parameters $F^{\star}$, $G^{\star}$ were randomly determined by following equations;
\begin{equation}
  F^{\star} = \exp\left[{\Delta t(A - A^{T})}\right],\
  \forall a \in A, a \sim \mathcal{N}(0, 1),
  \label{eqn:dynamics_F}
\end{equation}
\begin{equation}
  G^{\star} = \left(
  \begin{matrix}
  G_{c} \\
  O_{2, 2}
  \end{matrix} \right),
  G_{c} \in \mathbb{R}^{2 \times 2},
  \forall g \in G_{c}, g \sim \mathcal{N}(0, \Delta t), \label{eqn:dynamics_G}
\end{equation}
where $\exp[\cdot]$ indicates the matrix exponential and the time step size $\Delta t$ is 0.01.
Cost parameters $Q^{\star}$, $R^{\star}$ were deterministically given by;
\begin{equation}
	Q^{\star} = I_{4, 4} \cdot \Delta t, R^{\star} = I_{2, 2} \cdot \Delta t.
\end{equation}

Elements of a state vector $\mathbf{x}^{\star}_{t_0}$ were sampled from $\mathcal{N}(0, 1)$ and then the vector was input to LQR to generate a control sequence $\{\mathbf{u}^{\star}_{t_i}\}$ whose length was $N=200$.

\paragraph{PI-Net settings}
The internal dynamics parameters were initialized in the same manner as Eqs.~(\ref{eqn:dynamics_F}, \ref{eqn:dynamics_G}).
According to the internal cost parameters, all values were sampled from $\mathcal{N}(0, \Delta t)$.
The number of trajectories was $K=100$ and the number of \textit{PI-Net Kernel} recurrence was $U=200$.
We used the standard deviation $\sigma = 0.2$ to generate Gaussian noise $\delta \mathbf{u}$.

\begin{figure}[ht]
  \centering
  \includegraphics[width=0.5\textwidth]{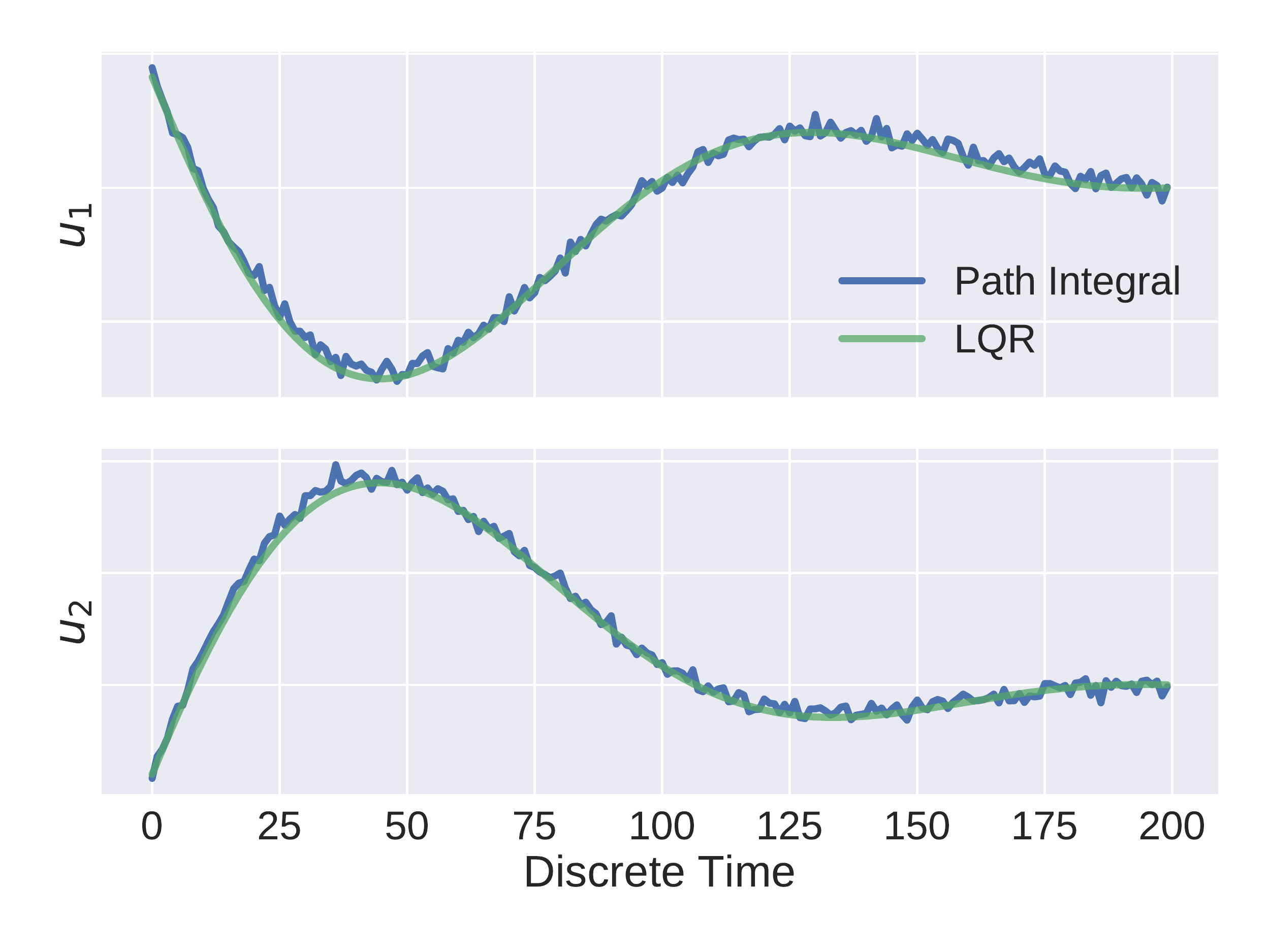}
  \caption{Control sequences} \label{fig:linear_ctrl_seq}
\end{figure}

\paragraph{Results}\
Fig.~\ref{fig:linear_ctrl_seq} exemplifies control sequences for an \textit{unseen} state input estimated by trained PI-Net and LQR.

\subsection{Pendulum Swing-Up}
\paragraph{Demonstrations}
We supposed that the system is dominated by the time-evolution: $\ddot{\theta} = -\sin(\theta) + k \cdot u$, where $k$ was set to be 0.5.
We discretized this continuous dynamics by using the forth-order Runge-Kutta method,
where the time step size was $\Delta t = 0.1$.

MPC simulations were conducted under the following conditions.
Initial pendulum-angles $\theta$ in trajectories were uniformly sampled from $[-\pi, \pi]$ at random.
Initial angular velocities were uniformly sampled from $[-1, 1]$.
iLQR output a control sequence whose length was $N=30$, and only the first value in the sequence was utilized for actuation at each time step.

\paragraph{PI-Net settings}
A neural dynamics model with one hidden layer (12 hidden nodes) and one output node was used to approximate $\ddot{\theta}$. $\dot{\theta}$ and $\theta$ were estimated by the Euler method utilizing their time derivatives.
The cost model is represented as $q(\theta, \dot{\theta}) = ||\mathsf{q}(\theta, \dot{\theta})||^{2}$, where $\mathsf{q}$ is a neural network with one hidden layer (12 hidden nodes) and 12 output nodes.
In the both neural networks, hyperbolic tangent is used as activation functions for hidden nodes.
Other PI-Net parameters were: $\sigma = 0.005$, $K=100$, $U=200$ and $N=30$.

\paragraph{VIN settings}
The VIN design for the pendulum task is described here.
In order to know the value iteration algorithm and the design methodology of VINs, the authors recommend to read Ref.~\cite{tamar2016value}.
The equation of reward propagation appeared in the algorithm, i.e., $\mathcal{Q}(s,a) \leftarrow \sum_{s'}\mathcal{V}(s')\mathcal{P}(s'|s,a) + \mathcal{R}(s,a)$, was represented by LCN or CNN layer, where  $\mathcal{Q}(s,a)$, $\mathcal{V}(s)$, $\mathcal{R}(s, a)$ were action-value map, value-map, and reward map, respectively. State-transition kernels $\mathcal{P}(s'|s,a)$ were represented by $7 \times 7$ kernels of LCN or CNN. This layer processes the input value map and creates a matrix $(\in \mathbb{R}^{31 \times 31 \times 7})$, then this matrix is summed up with the reward map to compute action-state map.
The attention module output selected action-state value, i.e.,~$\mathcal{Q}(s, \cdot) \in \mathbb{R}^{7}$ where $s$ corresponds input continuous states.
The reactive policy was a neural network with one hidden layer with 16 hidden nodes.
This neural network has 9D inputs (2D for states and 7D for output from the attention module) and 1D output for control. The activation function is rectified linear function.
The number of recurrence of VIN module is 50.

\end{document}